\documentclass[12pt]{article}

\usepackage[T2A]{fontenc}
\usepackage[utf8]{inputenc}
\usepackage{hyphenat}
\usepackage{xcolor}
\usepackage{hyperref}
\definecolor{linkcolor}{HTML}{000000}
\definecolor{citecolor}{HTML}{0B6B1A}
\definecolor{urlcolor}{HTML}{0B2A5B}
\hypersetup{pdfstartview=FitH, citecolor=citecolor, linkcolor=linkcolor, urlcolor=urlcolor, colorlinks=true}

\usepackage[pdftex]{graphicx}
\graphicspath{{.\\}}
\DeclareGraphicsExtensions{.pdf,.png,.jpg}

\usepackage{authblk}

\usepackage{array}
\usepackage{appendix}

\providecommand{\keywords}[1]{\vspace{10pt} \textit{\textbf{Keywords:} #1}}

\begin{document}

\title{Empirical investigations on WVA \\ structural issues}
\author[ ]{Alexey Kutalev$^1$ and Alisa Lapina$^2$}
\affil[1]{SberDevices, PJSC Sberbank, Moscow, Russia}
\affil[2]{Games department, VK, Moscow, Russia}
\affil[ ]{\it kutalev@gmail.com, ahm.alisa@gmail.com}
%\email{kutalev@gmail.com, ahm.alisa@gmail.com}

%\shortauthors{А. Куталев, А. Лапина}
 
%\received{25 August 2021}
%\revised{2 September 2021}
\date{\today}

\maketitle

\begin{abstract}

In this paper we want to present the results of empirical verification of some issues concerning the methods for overcoming catastrophic forgetting in neural networks. First, in the introduction, we will try to describe in detail the problem of catastrophic forgetting and methods for overcoming it for those who are not yet familiar with this topic. Then we will discuss the essence and limitations of the WVA method which we presented in previous papers. Further, we will touch upon the issues of applying the WVA method to gradients or optimization steps of weights, choosing the optimal attenuation function in this method, as well as choosing the optimal hyper-parameters of the method depending on the number of tasks in sequential training of neural networks.


\end{abstract}

\section{Introduction}

This section is intended to describe the problem of catastrophic forgetting of neural networks and actual methods for overcoming it for those who are not familiar with the topic at all or familiar by hearsay. The reader who better understands the problem can immediately proceed to the next section.

\subsection{Description of the problem}

So, the essence of the problem of catastrophic forgetting in neural networks is that when tasks A and B are trained sequentially, that is, when after training task A the neural network is trained to task B without using and accessing data of task A, the neural network in the process of training task B quickly loses the skills obtained when training to task A.
\begin{figure}[h]
\centering
% Areas of different tasks solutions in the NN parameter space
\includegraphics[width=1\textwidth]{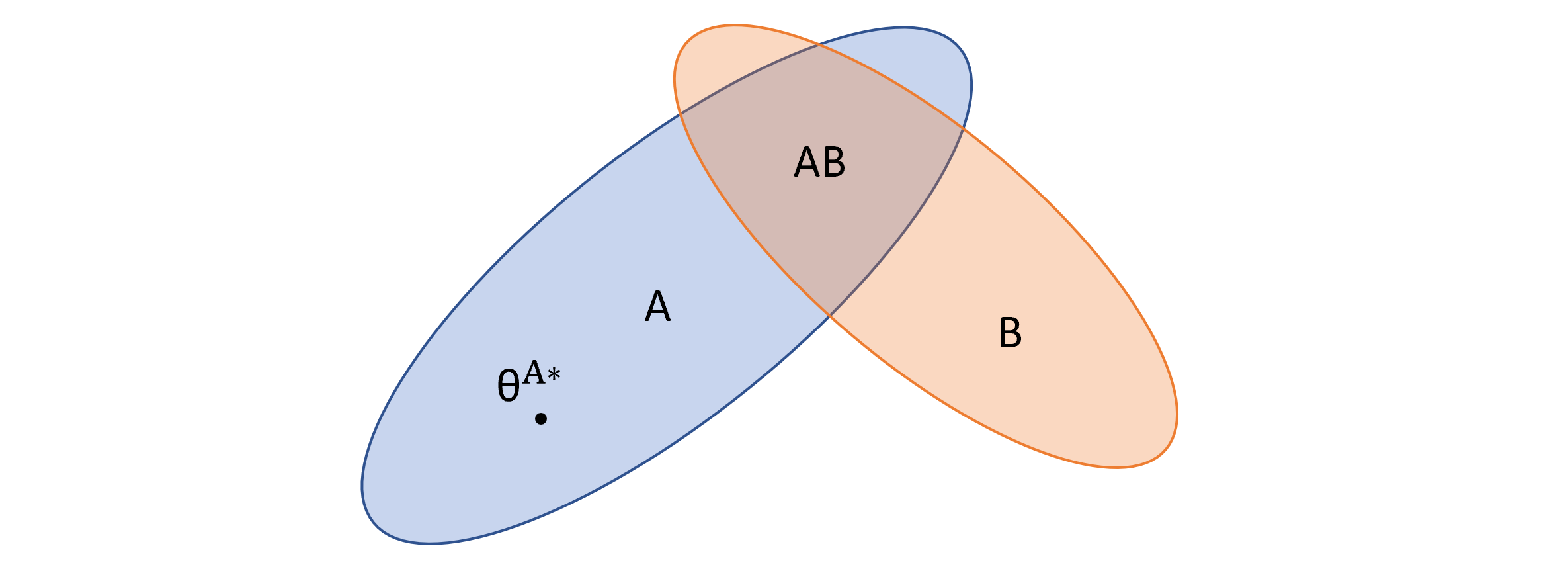}
\caption{\textit{Here A -- the area in the weight space of the neural network with low error on task A, and B -- the area with low error on task B.}}
\label{figure:1}
\end{figure}

Considerable efforts have been made for a long time to overcome catastrophic forgetting (see, for example, \cite{c1}-\cite{c4}), which, however, did not lead to significant overall results.

Great progress was made in 2017 when the Elastic Weight Consolidation (EWC) method was proposed by the DeepMind team \cite{c5}. This method is based on several assumptions. Firstly, it is assumed that in the space of weights of the neural network there is a region that is a solution for tasks that are already learned and for tasks in the queue for further learning (i.e., the areas in the space of weights that are solutions to each task have a non-empty intersection). Secondly, it is assumed that each weight of the neural network has its own importance for already learned tasks, and when training new tasks, the more it is necessary to keep each weight of neural network from change the more importance it has.

Whereas the first assumption requires only sufficient capacity of the neural network and that all tasks have no inconsistencies, for the justification of the second assumption we can propose the reasoning in the next subsection.

\subsection{Theoretical justification of the EWC method}

Let the loss function $L_D(\theta) = -\log p(D | \theta)$ (negative log loss is the most widely used loss function) be used to train a neural network with parameters (weights) $\theta$ on dataset $D$. Then, when training on data $D$, which is the union of two datasets $A$ and $B$, provided that $A$ and $B$ are independent, we have a loss function:
\begin{equation}
\label{equ1}
L_D(\theta) = -\log p(D | \theta) = -\log p(A | \theta) - \log p(B | \theta) = L_A(\theta) + L_B(\theta)
\end{equation}

Suppose we have already trained the neural network on the dataset $A$ to the required accuracy. That is, the convergence of some gradient method of training the neural network on the loss function $L_A(\theta)$ to the local minimum $\theta^*$ for the task $A$ is achieved. 

Convergence means that at the reached point $\theta^*$ the gradient of the loss function is zero or negligibly small. Then, if we represent the loss function $L_A$ in some neighbourhood of the solution point $\theta^*$ of task $A$ in the Taylor series with accuracy up to the second order terms, we have:
\begin{equation}
\label{equ2}
L_A(\theta) \approx L_A(\theta^*) + \sum_{i,j} H_{ij}(\theta^*) (\theta_i - \theta^*_i) (\theta_j - \theta^*_j),
\end{equation}
where $H_{ij}(\theta^*)$ is the Hesse matrix of the loss function at the point $\theta^*$. Since the point $\theta^*$ is the solution of the task $A$, that is, the loss function $L_A$ has a local minimum at the point $\theta^*$, the Hesse matrix $H_{ij}(\theta^*)$ is positive definite. 

The authors of \cite{c5} state that, according to \cite{c15} and \cite{c16}, the above-mentioned Hessian can be approximated by the form
\begin{equation}
\label{equ3}
\sum_{i,j} H_{ij}(\theta_i - \theta^*_i) (\theta_j - \theta^*_j) \approx \sum_i F_i (\theta_i - \theta^*_i)^2, 
\end{equation}
where $F_i(\theta^*) = \left( \frac{\partial \log p(A | \theta^* )}{\partial \theta_i}\right)^2$ are diagonal elements of the Fisher information matrix. As we can see, these elements depend only on the first derivatives of the loss function $L_A$, and therefore are easily computable.

Given \ref{equ2} and \ref{equ3}, we get that in the neighbourhood of the point $\theta^*$ the loss function $L_A(\theta)$ is approximated by the form $\sum_i F_i(\theta_i - \theta^*_i)^2$ up to a constant.

Further, following the logic of continual learning, we want to continue training the neural network on the task $B$ preserving the skill obtained by training on task $A$, that is to train the neural network on the data $D = A \cup B$. To do this, it is necessary to optimize the overall loss function $L_{D}(\theta)$. However, at the stage of training on task $B$ the data of task $A$ is no longer available. Therefore, in the formula \ref{equ1} we replace the component $L_A(\theta)$ with its approximation from \ref{equ2} and \ref{equ3}, and discard the constant $L_A(\theta^*)$, since it does not affect optimization:
\begin{equation}
\label{equ4}
L_{AB}(\theta) \approx L_B(\theta) + \frac{\lambda}{2}\sum_i F_i(\theta_i^*) (\theta_i - \theta^*_i)^2.
\end{equation}

Here the coefficient $\lambda$ was introduced, which determines the balance of contributions from the loss function $L_B$ and the approximation of the loss function $L_A$ to the total loss function $L_{AB}$. According to \cite{c5}, the form $F_i(\theta_i^*) (\theta_i - \theta^*_i)^2$ underestimates the Hessian from $L_A$. So, this underestimation can be compensated by fitting $\lambda$.

Thus, the training of the neural network on the dataset $B$ is reduced to the optimization of the loss function in the form of \ref{equ4}, and due to its second part, the more each weight $\theta_i$ of the neural network resists changes the more importance it has. Where importance equals to $F_i$, calculated in the point $\theta^*$. Q.E.D.

\subsection{Further development of the EWC method}

Later, in the papers \cite{c7}, \cite{c8} and \cite{c10} alternative ways of calculating the importance of weights were proposed. Moreover, experiments in \cite{c8} and \cite{c11} show that the MAS method from \cite{c8} gives the most optimal values of the importance of weights for preserving previous skills when used instead of $F_i$ in the loss function \ref{equ4} for continual learning.

To generalize the method to the case of sequential learning of several tasks, the authors of \cite{c5} propose to add components to the overall loss function approximating the loss function for each already learned task $A$, $B$, ... , $J$:
\begin{equation}
\label{equ5}
L(\theta) = L_K(\theta) + \frac{\lambda_A}{2}\sum_i F_i(\theta_i^{A*}) (\theta_i - \theta^{A*}_i)^2 + 
\end{equation}
$$
+ \frac{\lambda_B}{2}\sum_i F_i(\theta_i^{B*}) (\theta_i - \theta^{B*}_i)^2 + ... + \frac{\lambda_J}{2}\sum_i F_i(\theta_i^{J*}) (\theta_i - \theta^{J*}_i)^2
$$

However, as it was showed by Husz\'ar in \cite{c6}, to train each subsequent task $K$ it is more correct to sum up all the importances of each weight on already learned tasks $F_i(\theta^{A*}) + F_i(\theta^{B*}) + ... + F_i(\theta^{J*})$ and use $\theta^{J*}$ -- network weights after the last learned task $J$ as the point of attraction:
\begin{equation}
\label{equ6}
L(\theta) = L_K(\theta) + \frac{\lambda}{2}\sum_i \left[ F_i(\theta_i^{A*}) + ... + F_i(\theta_i^{J*}) \right] (\theta_i - \theta^{J*}_i)^2,
\end{equation}
that significantly reduces the computational cost compared to the formula \ref{equ5}.

To balance between preserving the skills of the already learned tasks and learning the current task, $F_i(\theta_i^*)$ can be summed up in the formula \ref{equ6} with coefficients proportional to the importances of the learned tasks. A similar mechanism was used in \cite{c9} to lessen the importance of tasks as they move further away from the current learning, and was called $Online EWC$, which is often referred to in modern papers on catastrophic forgetting.

\subsection{Features of EWC practical use}

As we mentioned in our paper \cite{c11}, when using the EWC method for convolutional, recurrent or self-attention based networks, the importance of some weights may become extremely large, and this leads to undesirable effects when applying EWC to training neural networks, such as "gradient explosion" or accuracy loss.

To fight the case of "gradient explosion" produced by extremely large values of weight importances, it has been proposed to use a modified loss function in the following form:
$$
L(\theta) \approx L_T(\theta) + \frac{\lambda}{2}\sum_i \frac{\Omega_i}{\alpha\lambda \Omega_i + 1} (\theta_i - \theta^*_i)^2,
$$
where $L_T$ is the loss function for the current task $T$, $\Omega_i$ is the accumulated importance of $i$-th weight on all previous tasks, $\alpha$ is the learning rate of the neural network. This modification allows for the term that approximating Hessian to contribute to the weight optimization step without exceeding the distance between the weight $\theta_i$ and its consolidated value $\theta^*_i$.

The accuracy loss problem is the situation when the Hessian approximation term makes so large a contribution by the norm to the gradient of loss function which is several orders larger than the contribution of the gradient of loss function for the current task $L_T(\theta)$. To overcome this problem we suggested in our paper \cite{c11} to perform "gradient clipping" separately for the gradient of the loss function $L_T(\theta)$ of the current task and the gradient of the term approximating the loss function of previous tasks, and then to sum up the results in a single vector-gradient for use by the optimization algorithm.

\subsection{Inaccuracies and open-ended questions of the method}

Finally, it is necessary to mention the inaccuracies and limitations of the EWC method.

First, there is no theoretical justification for the fact that the subspace of solutions of the task in the space of neural network weights is connected and continuous. Therefore, it is not obvious that by moving inside the solution space of task $A$ using the optimization algorithm the solution space of task $B$ can be reached.

Second, it is known (see \cite{c17}, \cite{c1}) that for neural networks in a small neighbourhood of the solution $\theta^*$ of task $A$ there can be points $\theta'$ for which the loss function $L_A(\theta')$ differs very significantly from the optimal value $L_A(\theta^*)$, which puts into question the correctness of its approximation using the quadratic form.

Third, the above justification of EWC is valid for neural network classifiers, i.e. networks with a loss function of the form $L_D(\theta) = \log p(D | \theta)$, which limits the scope of the method.

Fourth, the justification uses the independence of the datasets in the sequential tasks $A$ and $B$, which is most often not the case.

Finally, fifth, EWC does not restrict in any way the change in weights at the first optimization step of training the next task, since the first step is taken from the point $\theta^*$, and at this point both the term approximating the Hessian and its gradient are equal to zero. It is true that the subsequent steps correct the situation in most cases, i.e. they return the important weights closer to their consolidated values in proportion to their importance.

However, in spite of all the listed inaccuracies, the EWC method can quite successfully fight catastrophic forgetting in practice.

\section{Weight velocity attenuation method (WVA)}

As a simplified alternative to EWC, the Weight Velocity Attenuation (WVA) method was proposed in \cite{c10}. The simplification with respect to EWC is that it is not required to save the weights of the neural network $\theta^*$ after each or only after the last task -- it is required only to accumulate and store the importance values of the weights.

\begin{figure}[h]
\centering
\includegraphics[width=1\textwidth]{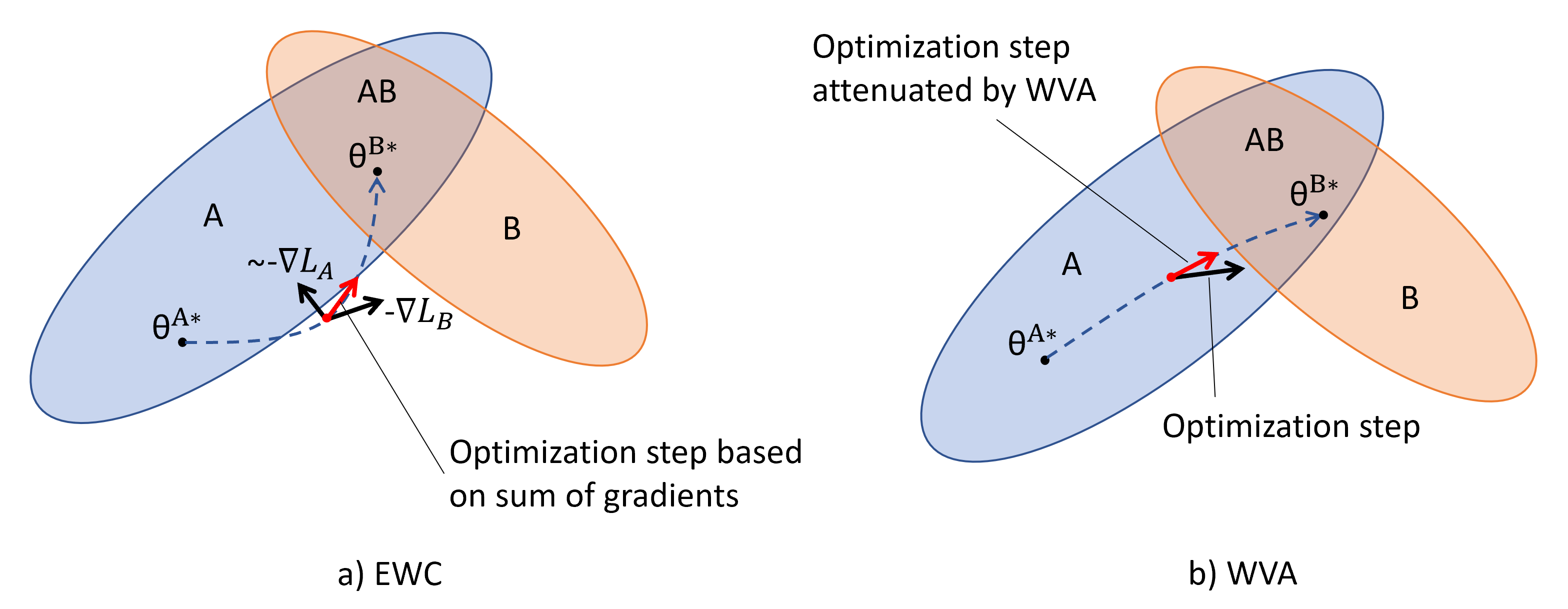}
\caption{\textit{Illustration of training a neural network for task B from a point that is a solution to task A using EWC (left) and WVA (right) methods.}}
\label{figure:2}
\end{figure}
The essence of the method is that instead of keeping the weights drawn to the anchor points $\theta^*_i$ the optimization step for each parameter of the neural network is attenuated in proportion to the importance of that parameter. For example, this can be done using a hyperbolic function:
\begin{equation}
\label{equ7}
\theta^{t+1}_i = \theta^{t}_i + \frac{ \Delta \theta^{t}_i } { \lambda \Omega_i + 1},
\end{equation}
where $\Delta\theta^t_i$ -- optimization step, generated by the gradient method, $\Omega_i$ -- accumulated importance of $i$-th weight after several previous tasks learned by neural network, $\lambda$ -- common factor for the whole neural network, which determines how much the network tends to keep the learned skills against learning the current one.

Here we try to justify -- why WVA allows us to preserve skills during continual learning. To do this, consider the loss function used in the EWC method for learning $T$ task after learning several previous tasks:
\begin{equation}
\label{equ8}
L(\theta) = L_T(\theta) + \frac{\lambda}{2}\sum_i \Omega_i (\theta_i - \theta^*_i)^2,
\end{equation}
here $L_T(\theta)$ is the loss function on the dataset of task $T$, $\Omega_i$ is the accumulated importance of $i$-th weight after learning on several previous tasks, $\lambda$ is the balance between keeping the skills learned on previous tasks and learning the current task $T$.

As we know, the quadratic regularizer $\sum_i \Omega_i (\theta_i - \theta^*_i)^2$ in the formula \ref{equ8} from the probabilistic point of view corresponds to the normal distribution $p(\theta|D)$ (here $D$ is the union of all task datasets, already learned by the network) of parameters $\theta_i$ with a mean of $\theta^*_i$ and standard deviations $\sigma_i = \sqrt\frac{1}{2 \Omega_i}$. Thus, in order to preserve the skills learned on $D$, each parameter $\theta_i$ must remain within some (confidence) interval defined for it. Given this, we see that the formula \ref{equ7} performs this function: it slows down changes in parameters with small variance (that is, with large importance $\Omega_i$) and, conversely, allows parameters with large variance to change more.

Note that, unlike EWC, WVA still allows the optimizer to take parameters from their previous values indefinitely far beyond the confidence interval if the current task is trained for a sufficient number of epochs.
Therefore, WVA can be useful in case it is not necessary to exactly preserve the skills, but it is useful to keep the internal representations learned in the network on previous tasks.

During the development of the WVA method, we had to answer several questions, which we will discuss in more detail below. For their experimental verification, we used a three-layer neural network with two fully connected layers for 300 and 150 perceptrons with leakyReLU activation and an output softmax layer for 10 perceptrons. In each experiment, training was performed sequentially on 10 datasets obtained from the MNIST dataset by random permutation of inputs (permuted MNIST, see \cite{c4}, \cite{c18}). Training on each of the datasets was carried out for 4 epochs with a minibatch size of 100. The learning rate was chosen to be 0.001 for Adam and 0.2 for SGD. The importances of the weights were calculated by the method of the total absolute signal (see \cite{c10}). We also performed similar experiments with the importances of weights calculated based on the Fisher information matrix (see \cite{c5}) and got a similar picture. Therefore, below we present only the results of experiments with importances based on the total absolute signal.

\subsection{What is the attenuation applied to -- the gradient of the loss function or the optimization step?}

In the case of using simple stochastic gradient descent (SGD) as a neural network training method, the gradient of the loss function multiplied by a constant is used as an optimization step. Therefore, for SGD there is no difference to apply attenuation to the gradient or optimization step.
\begin{figure}[h]
\centering
\includegraphics[width=1\textwidth]{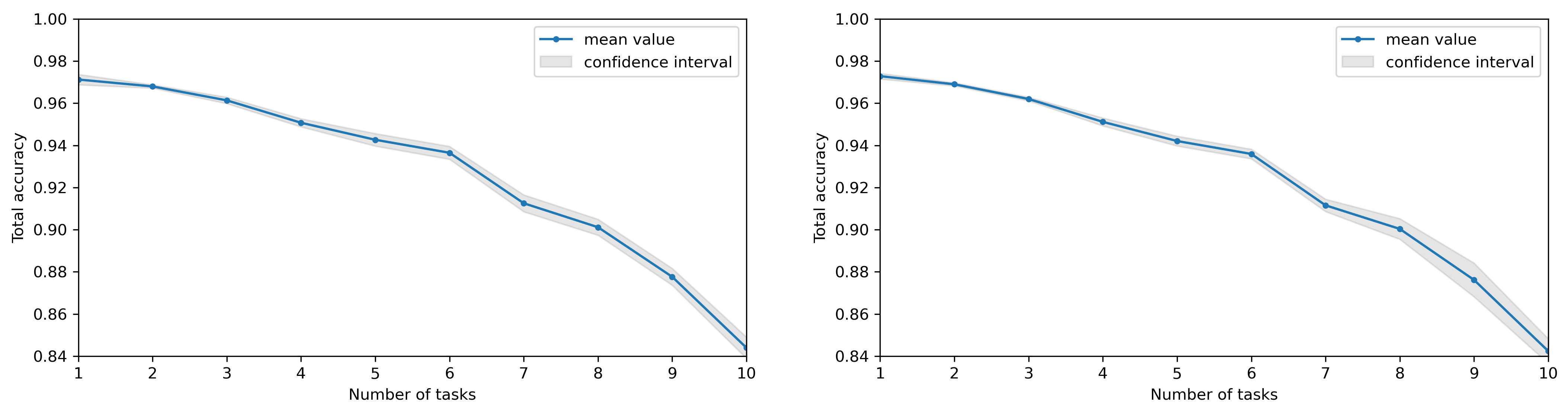}
\caption{\textit{Plots of achievable accuracy after training on 10 datasets by SGD optimizer: on the left plot WVA is applied to the gradient of the loss function, on the right plot -- to the optimization step. The figures are absolutely identical.}}
\label{figure:3}
\end{figure}

When using an optimizer with more complex logic (e.g., SGD with momentum, AdaGrad, AdaMax, RMSprop, etc.), it is intuitively clear that attenuation should be applied to the optimization step created by the optimizer from the gradient, because the optimizer can change the proportionality between the gradient and optimization step following its own logic and, for example, greatly increase the optimization step for the weight with high importance. So when you apply WVA to the gradient, the optimization steps on the important weights can become large again. However, when WVA is applied to the optimization steps, such effect does not occur. 

\begin{figure}[h]
\centering
\includegraphics[width=1\textwidth]{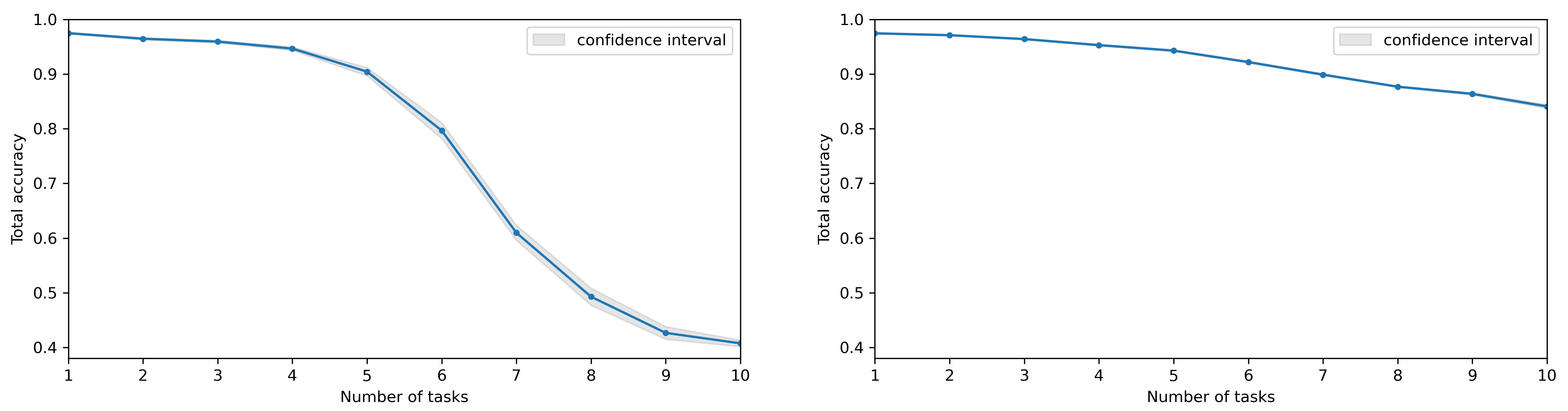}
\caption{\textit{Plots of achievable accuracy after training on 10 datasets by Adam optimizer: on the left plot WVA is applied to the gradient of loss function, on the right plot -- to the optimization step.}}
\label{figure:4}
\end{figure}

To confirm this reasoning, we performed an appropriate experiment. The optimal (providing the highest average accuracy on all learned datasets) hyper-parameter $\lambda$ was obtained for each of the cases by a grid search. The results of the experiment can be seen on the figure \ref{figure:4}. It is obvious how much more the accuracy degraded on all learned datasets when attenuation was applied to the gradient of the loss function.

\subsection{What should be the attenuation function? Let's compare hyperbolic and exponential attenuation}

WVA-attenuation on optimization step can be implemented in various ways. We considered the methods used in reinforcement learning to discount the reward (see \cite{c12}, \cite{c13}, \cite{c14}). Attenuation using a hyperbolic function has already been presented in the formula \ref{equ7}, and exponential attenuation is implemented by the following formula:
\begin{equation}
\label{equ9}
\theta^{t+1}_i = \theta^{t}_i + e^{-\lambda \Omega_i} \Delta \theta^{t}_i.
\end{equation}

\begin{figure}[h]
\centering
\includegraphics[width=1\textwidth]{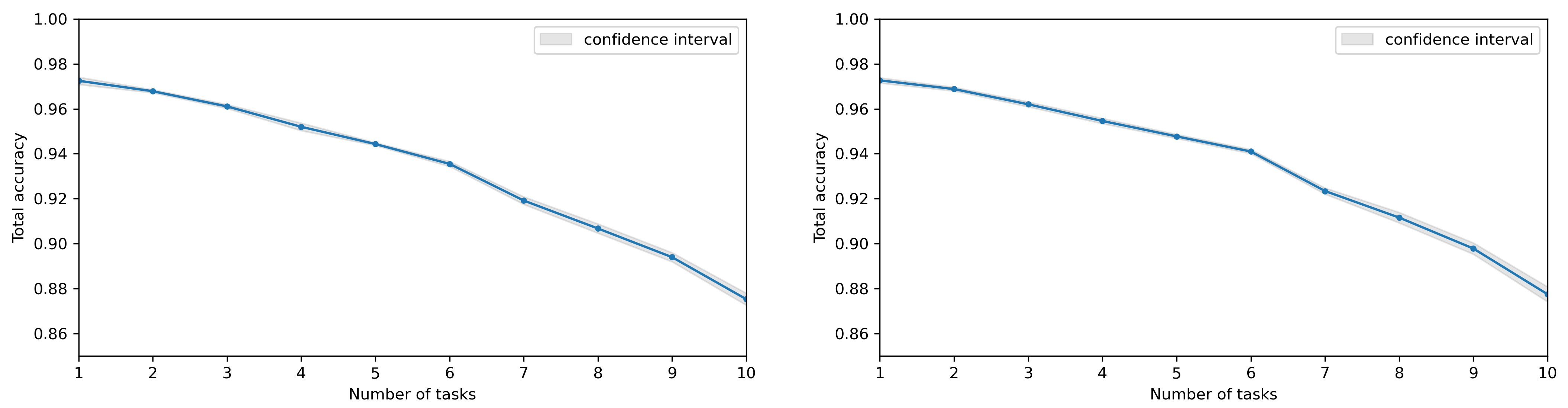}
\caption{\textit{Plots of achievable accuracy after training on 10 datasets by SGD optimizer: on the left plot WVA is performed by a hyperbolic function, on the right plot -- by an exponent.}}
\label{figure:5}
\end{figure}

\begin{figure}[h]
\centering
\includegraphics[width=1\textwidth]{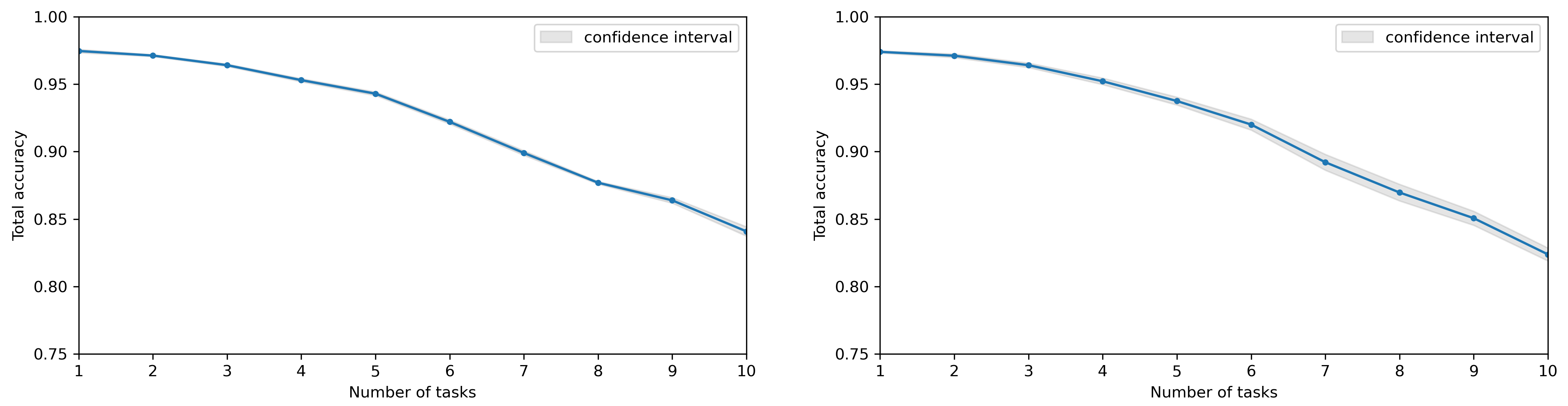}
\caption{\textit{Plots of achievable accuracy after training on 10 datasets with Adam optimizer: on the left plot WVA is performed by hyperbolic function, on the right plot -- by exponent.}}
\label{figure:6}
\end{figure}

Similar to the previous case, we found the optimal $\lambda$ by grid search (see \cite{c11} for methodology) for the attenuation by hyperbolic function and by exponent and using SGD and Adam optimizers. The results can be seen in the figures \ref{figure:5} and \ref{figure:6}.

In the case of using SGD, the exponential attenuation was a bit better than the hyperbolic; in the case of Adam, it was a bit worse. We see that the difference between the attenuation types is insignificant at the optimal $\lambda$. Thus, no unambiguous conclusions can be drawn about the advantages of any of these functions.

\subsection{How does the optimal value of the hyper-parameter $\lambda$ change depending on the number of tasks in continual learning?}

Earlier, when making a comparison, we used a grid search for the optimal hyper-parameter $\lambda$ for each configuration. Here we present plots of the surfaces of the accuracy dependence on $\lambda$ and the number of datasets in the sequential training obtained as a result of this search.
\begin{figure}[h]
\centering
\includegraphics[width=1\textwidth]{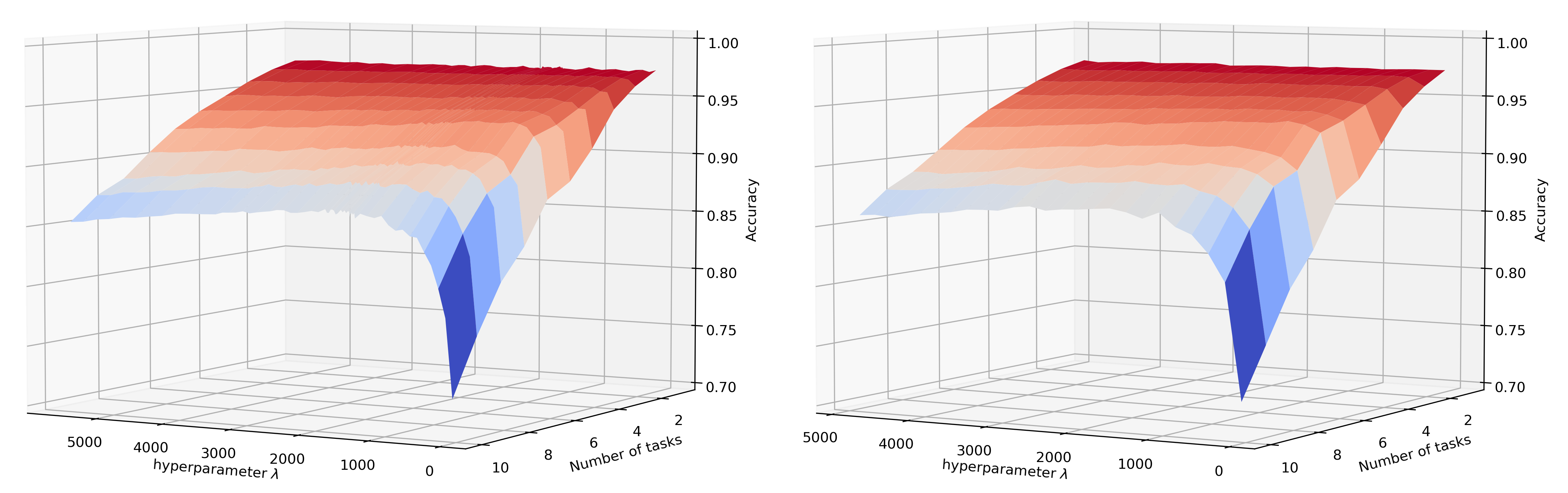}
\caption{\textit{Surfaces of average accuracy as a function of $\lambda$ and the number of learned datasets obtained using SGD optimizer: on the left plot WVA is applied to the gradient of loss function, on the right plot -- to the optimization step.}}
\label{figure:7}
\end{figure}

\begin{figure}[h]
\centering
\includegraphics[width=1\textwidth]{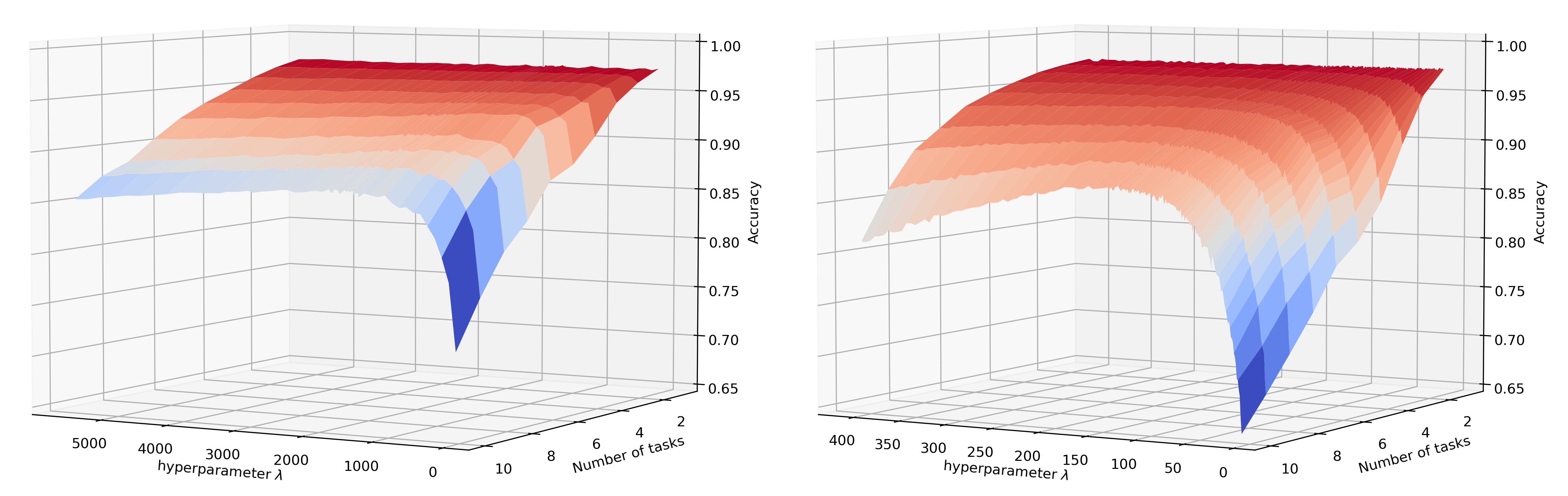}
\caption{\textit{Surfaces of average accuracy as a function of $\lambda$ and the number of learned datasets obtained using SGD optimizer: on the left plot WVA is performed by a hyperbolic function, on the right plot -- by an exponent.}}
\label{figure:8}
\end{figure}

\begin{figure}[h]
\centering
\includegraphics[width=1\textwidth]{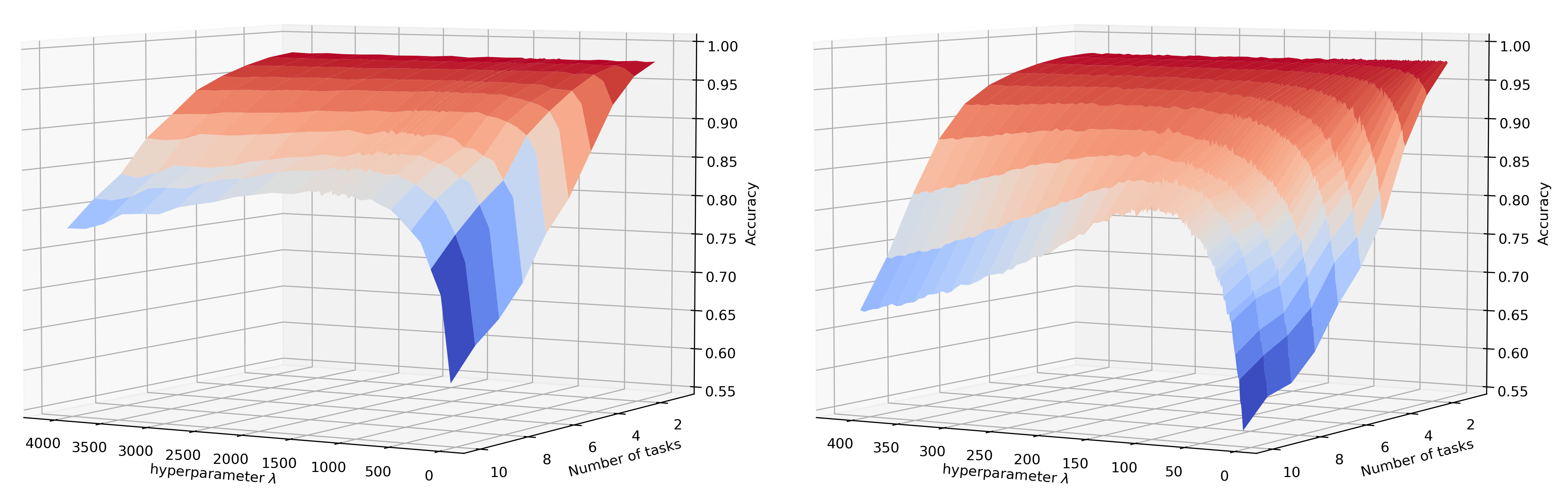}
\caption{\textit{Surfaces of average accuracy as a function of $\lambda$ and the number of learned datasets obtained using Adam optimizer: on the left plot attenuation is produced by a hyperbolic function, on the right plot -- by an exponent.}}
\label{figure:9}
\end{figure}

On these surfaces on the plots \ref{figure:7}, \ref{figure:8}, \ref{figure:9} one can observe that the optimal (with maximum accuracy) value of $\lambda$ is mostly independent of the number of learned datasets. Thus, to save resources, one can perform the grid search of $\lambda$ on a small number of sequential datasets and expect the same $\lambda$ value would be optimal when training on a longer sequence if the datasets in this sequence have a similar structure.

On the plots, we can also see that when the hyper-parameter $\lambda$ moves away from the optimal value, the accuracy obtained by exponential attenuation degrades much more than obtained by hyperbolic attenuation.

\section{Conclusion}

The purpose of this paper was to share the results of our experiments that we conducted creating the Weights Velocity Attenuation method (WVA). Without claiming generality or complete mathematical rigor, we can still formulate some conclusions and assumptions about the use of WVA. 

For example, we confirmed that attenuation should be applied to the weight optimization step rather than to the gradient of the loss function. We also found that the exponential attenuation function gives about the same quality as the hyperbolic one at the optimal value of the hyper-parameter $\lambda$. But when moving away from the optimal $\lambda$ the exponential attenuation shows worse performance, so the optimal choice would be the hyperbolic one.

Based on observations during the calculation of the hyper-parameter $\lambda$, which is responsible for the balance between skill preserving and further learning, it can be assumed that the optimal value of $\lambda$ is independent of the number of tasks in the continual learning, and thus can be found by grid search on a short sequence of tasks, and practically used on a longer sequence.

%-----------------------


\newpage

\begin{thebibliography}{99}

\bibitem{c1} McCloskey M, Cohen NJ (1989) Catastrophic interference in connectionist networks: The sequential learning problem. \textit{The Psychology of Learning and Motivation}, ed GH Bower (Academic, New York), Vol 24, pp 109-165.

\bibitem{c2} McClelland JL, McNaughton BL, O'Reilly RC (1995) Why there are complementary learning systems in the hippocampus and neocortex: Insights from the successes and failures of connectionist models of learning and memory. \textit{Psychological Review}, Vol 102(3), pp 419-457.

\bibitem{c3} French RM (1999) Catastrophic forgetting in connectionist networks. \textit{Trends in Cognitive Sciences}, Vol 3(4), pp 128-135.

\bibitem{c4} Goodfellow IJ, Mirza M, Xiao D, Courville A, Bengio Y (2015) An empirical investigation of catastrophic forgetting in gradient-based neural networks. \textit{arXiv}:1312.6211.

\bibitem{c5} Kirkpatrick J, Pascanu R, Rabinowitz N, Veness J, Desjardins G, Rusu AA, Milan K, Quan J, Ramalho T, Grabska-Barwinska A, Hassabis D, Clopath C, Kumaran D, Hadsell R (2017) Overcoming catastrophic forgetting in neural networks. \textit{Proceeding of the National Academy of Science}, Vol. 114(13), pp 3521–3526. DOI: 10.1073/pnas.1611835114

\bibitem{c6} Husz\'ar F (2018) Note on the quadratic penalties in elastic weight consolidation. \textit{Proceeding of the National Academy of Science}, Vol. 115(11), pp 2496–2497. DOI: 10.1073/pnas.1717042115

\bibitem{c20} Kirkpatrick J, Pascanu R, Rabinowitz N, Veness J, Desjardins G, Rusu AA, Milan K, Quan J, Ramalho T, Grabska-Barwinska A, Hassabis D, Clopath C, Kumaran D, Hadsell R (2018) Reply to Husz\'ar: The elastic weight consolidation penalty is empirically valid \textit{Proceeding of the National Academy of Science} Vol. 115(11). DOI: 10.1073/pnas.1800157115

\bibitem{c7} Zenke F, Poole B, and Ganguli S (2017) Continual learning through synaptic intelligence. \textit{Proceedings of the 34th International Conference on Machine Learning (ICML)}

\bibitem{c8} Aljundi R, Babiloni F, Elhoseiny M, Rohrbach M and Tuytelaars T (2018) Memory aware synapses: Learning what (not) to forget. \textit{The European Conference on Computer Vision (ECCV)}, September 2018.

\bibitem{c9} Schwarz J, Czarnecki W, Luketina J, Grabska-Barwinska A, Teh YW, Pascanu R, Hadsell R (2018, July). Progress \& compress: A scalable framework for continual learning. \textit{In International Conference on Machine Learning }, pp. 4528-4537. PMLR.

\bibitem{c10} Kutalev A (2020) Natural Way to Overcome Catastrophic Forgetting in Neural Networks. \textit{Modern Information Technologies and IT-Education}, Vol. 16(2), pp 331-337. DOI: 10.25559/SITITO.16.202002.331-337.

\bibitem{c11} Kutalev A, Lapina A (2021) Stabilizing Elastic Weight Consolidation method in practical ML tasks and using weight importances for neural network pruning. \textit{Modern Information Technologies and IT-Education}, Vol. 17(2), pp 345-354. DOI: 10.25559/SITITO.17.202102.345-354

\bibitem{c12} Ainslie G, Haslam N (1992). Hyperbolic Discounting. In: G. Loewenstein and J. Elster, Eds., \textit{Choice Over Time}, Russell Sage Foundation Publications, New York, 1992, pp. 57-92.

\bibitem{c13} Green L, Myerson J (1996). Exponential versus hyperbolic discounting of delayed outcomes: Risk and waiting time. \textit{American Zoologist}, Vol. 36(4), pp 496-505.

\bibitem{c14} Fedus W, Gelada C, Bengio Y, Bellemare MG, Larochelle H (2019) Hyperbolic discounting and learning over multiple horizons. \textit{arXiv}:1902.06865.

\bibitem{c15} MacKay DJ (1992) A practical Bayesian framework for backpropagation networks. \textit{Neural Computation}, Vol. 4(3), pp 448–472.

\bibitem{c16} Pascanu R, Bengio Y (2013) Revisiting natural gradient for deep networks. \textit{arXiv}:1301.3584

\bibitem{c17} Huang WR, Emam Z, Goldblum M, Fowl L, Terry JK, Huang F, Goldstein T (2019) Understanding Generalization through Visualizations. \textit{arXiv}:1906.03291

\bibitem{c18} Srivastava RK, Masci J, Kazerounian S, Gomez F, Schmidhuber J (2013) Compete to compute.  \textit{Advances in Neural Information Processing Systems 26}, eds Burges CJC, Bottou L, Welling M, Ghahramani Z, Weinberg KQ (Curran Assoc, Red Hook, NY), Vol 26, pp 2310–2318.

\bibitem{c19} Thangarasa V, Miconi T, Taylor GW (2020) Enabling Continual Learning with Differentiable Hebbian Plasticity \textit{arXiv}:2006.16558

\bibitem{c21} Miconi T, Stanley KO, Clune J (2018) Differentiable plasticity: training plastic neural networks with backpropagation. \textit{Proceedings of the 35th International Conference on Machine Learning}, PMLR 80:3559-3568.

\bibitem{c22} Gupta S, Singh P, Chang K, Qu L, Aggarwal M, Arun N, Vaswani A, Raghavan S, Agarwal V, Gidwani M, Hoebel K, Patel J, Lu C, Bridge CP, Rubin DL, Kalpathy-Cramer J (2021) Addressing catastrophic forgetting for medical domain expansion. \textit{arXiv}:2103.13511.

\bibitem{c23} Zenke F, Gerstner W, Ganguli S (2017) The temporal paradox of hebbian learning and homeostatic plasticity. \textit{Current Opinion in Neurobiology}. Vol. 43, pp 166-176. DOI: 10.1016/j.conb.2017.03.015

\bibitem{c24} van Garderen K, van der Voort S, Incekara F, Smits M, Klein S (2019) Towards continuous learning for glioma segmentation with
elastic weight consolidation. \textit{arXiv}:1909.11479

\bibitem{c25} Madasu A, Vijjini AR (2020) Sequential Domain Adaptation through Elastic Weight Consolidation for Sentiment Analysis \textit{arXiv}:2007.01189

\end{thebibliography}
\end{document}